\documentclass[11pt,a4paper]{article}
\usepackage[hyperref]{stylesheet/emnlp2018}
\usepackage{times}
\usepackage{latexsym}

\usepackage{url}

\aclfinalcopy 
\def\aclpaperid{1217} 

\usepackage{mathtools}
\usepackage{multirow}
\usepackage{amsmath}
\usepackage{amssymb}
\usepackage{booktabs}
\newcommand{\textsuper}[1]{$^\text{#1}$}
\usepackage{rotating}

\newcommand\blfootnote[1]{%
  \begingroup
  \renewcommand\thefootnote{}\footnote{#1}%
  \addtocounter{footnote}{-1}%
  \endgroup
}

\title{Modeling Empathy and Distress in Reaction to News Stories}

\author{
Sven Buechel \textsuper{*\dag3}
\hspace{.3cm} 
Anneke Buffone \textsuper{*1}
\hspace{.3cm} 
Barry Slaff \textsuper{1}
\hspace{.3cm}
Lyle Ungar \textsuper{1,2}  
\hspace{.3cm}
Jo\~{a}o Sedoc \textsuper{1,2}\vspace{.3cm}\\
\textsuper{1} Positive Psychology Center, University of Pennsylvania\\
\textsuper{2} Computer \& Information Science, University of Pennsylvania\\
\textsuper{3} JULIE Lab, Friedrich-Schiller-Universit\"at Jena \vspace{.3cm}\\
\url{https://wwbp.org} \textsuper{1,2} \hspace*{.5cm} \url{https://julielab.de} \textsuper{3}
}

\date{}

\begin{document}
\maketitle
\blfootnote{* These authors contributed equally to this work. Anneke Buffone designed and supervised the crowdsourcing task and the survey described in Section \ref{sec:corpus}, and provided psychological background knowledge. Sven Buechel was responsible for corpus creation, data analysis, and modeling. The technical set-up of the crowdsourcing task and the survey was done jointly by both first authors.}
\blfootnote{\dag Work conducted while being at the University of Pennsylvania.}
\begin{abstract}
Computational detection and understanding of empathy is an important factor in advancing human-computer interaction. Yet to date, text-based empathy prediction has the following major limitations: It underestimates the psychological complexity of the phenomenon, adheres to a weak notion of ground truth where empathic states are ascribed by third parties, and lacks a shared corpus. In contrast, this contribution presents the first publicly available gold standard for empathy prediction. It is constructed using a novel annotation methodology which reliably captures empathy assessments by the writer of a statement using multi-item scales. This is also the first computational work distinguishing between multiple forms of empathy, empathic concern, and personal distress, as recognized throughout psychology. Finally, we present experimental results for three different predictive models, of which a CNN performs the best.
\end{abstract}

\section{Introduction}
\label{sec:intro}
 
Over two decades after the seminal work by \newcite{Picard97} the quest of \textit{Affective Computing}, to ease the interaction with computers by giving them a sense of how emotions shape our perception and behavior, is still far from being fulfilled. Undoubtedly, major progress has been made in NLP, with sentiment analysis being one of the most vivid and productive areas in recent years \cite{Liu15}.

However, the vast majority of contributions has focused on \textit{polarity prediction}, typically only distinguishing between positive and negative feeling or evaluation, usually in social media postings or product reviews \cite{Rosenthal17,Socher13}. Only very recently, researchers started exploring more sophisticated models of human emotion on a larger scale \cite{Wang16acl,Abdul17acl, Mohammad17wassa, Buechel17eacl, Buechel18coling, Buechel18naacl}. Yet such approaches, often rooted in psychological theory, also turned out to be more challenging in respect to annotation and modeling \cite{Strapparava07}.
%

Surprisingly, one of the most valuable affective phenomena for improving human-machine interaction has received surprisingly little attention: \textit{Empathy}. Prior work focused mostly on {\it spoken dialogue}, commonly addressing conversational agents, psychological interventions, or call center applications \cite{Mcquiggan2007,Fung16,Perez-Rosas17,Alam17}. 

In contrast, to the best of our knowledge, only three contributions \cite{Xiao12,Gibson15,Khanpour17} previously addressed {\it text-based} empathy prediction\footnote{
    Psychological studies commonly distinguish between \textit{state} and \textit{trait} empathy. While the former construct describes the amount of empathy a person experiences as a direct result of encountering a given stimulus, the latter refers to how empathetic one is on average and across situations. This studies exclusively addresses \textit{state empathy}. For a contribution addressing \textit{trait empathy} from an NLP perspective, see \newcite{Abdul2017icwsm}.

} 
(see Section \ref{sec:modeling} for details). Yet, all of them are limited in three ways: (a) neither of their corpora are available leaving the NLP community without shared data, (b) empathy ratings were provided by others than the one actually experiencing it which qualifies only as a weak form of ground truth, and (c) their notion of empathy is quite basic, falling short of current and past theory.

In this contribution we present the first publicly available gold standard for text-based empathy prediction.
It is constructed using a novel annotation methodology which reliably captures empathy assessments via multi-item scales. The corpus as well as our work as a whole is also unique in being---to the best of our knowledge---the first computational approach differentiating \textit{multiple types of empathy}, empathic concern and personal distress, a distinction well recognized throughout psychology and other disciplines.\footnote{Data and code are  available at: \url{https://github.com/wwbp/empathic_reactions}}

\begin{table*}[t]
    \centering
    \small
    \begin{tabular}{p{.2cm}p{.2cm}p{.2cm}p{13.6cm}}
    \toprule
    & {\bf E} & {\bf D} & {\bf Message}\\
    \midrule
    (1) & 4.8     & 3.1 & {\it I'm sorry to hear that about Dakota's parents. Even when you are adult it must be hard to see your parents splitting up. No one wants that to happen and it's unfortunate that her parents couldn't work it out. I hope they are able to still remain civil around the kids and family. Just because it didn't work romantically doesn't mean it won't work at all.}\\
    \midrule
    (2) & 4.0    & 5.5 & {\it Here's an article about crazed person who murdered two unfortunate women overseas. Life is crazy. I can't imagine what the families are going through. Having to go to or being forced into sex work is bad enough, but for it to end like this is just sad. It feels like there's no place safe in this world to be a woman sometimes.} \\
    \midrule
    (3) & 1.0 & 1.3 & {\it I just read an article about some chowder-head who used a hammer and a pick ax to destroy Donald Trump's star on the Hollywood walk of fame. Wow, what a great protest. You sure showed him. Good job. Lol, can you believe this garbage? Who has such a hollow and pathetic life that they don't have anything better to do with their time than commit petty vandalism because they dislike some politician? What a dingus.}
    \\
    \bottomrule
    \end{tabular}
    \vspace*{-3pt}
    \caption{Illustrative examples from our newly created gold standard with ratings for empathy ({\bf E}) and distress ({\bf D}).}
    \label{tab:examples}
    \vspace*{-6pt}
\end{table*}
\section{Corpus Design and Methodology}
\label{sec:corpus}


\textbf{Background. }
 Most psychological theories of empathic states are focused on reactions to negative  rather than positive events. Empathy for positive events remains less well understood and is thought to be regulated differently \cite{morelli2015emerging}. Thus we focus on empathetic reactions to 
 need or suffering. 
 Despite the fact that everyone has an immediate, implicit understanding of empathy, research has been vastly inconsistent in its definition and operationalization  \cite{cuff2016empathy}. 
 There is agreement, however, that there are multiple forms of empathy (see below).  
 %
 %
 The by far most widely cited state empathy scale 
 is Batson's Empathic Concern -- Personal Distress Scale \cite{batson1987distress}, henceforth {\it empathy} and {\it distress}.  
 %
 
 Distress is a self-focused, negative affective state that occurs when one feels upset due to witnessing an entity's suffering or need, potentially via ``catching'' the suffering target's negative emotions. 
Empathy is a warm, tender, and compassionate feeling for a suffering target. It is other-focused, retains self-other separation, and is marked by relatively more positive affect 
\cite{batson1991evidence,goetz2010compassion,Mikulincer10,sober1997unto}.




\textbf{Selection of News Stories. }
Two research interns  (psychology undergraduates) collected a total of 418 articles from popular online news platforms, selected to likely evoke empathic reactions, after being briefed on the goal and background of this study. 
These articles were then used to elicit empathic responses in participants.

\textbf{Acquiring Text and Ratings. }
The corpus acquisition was set up as a crowdsourcing task on \texttt{MTurk.com} pointing to a  \texttt{Qualtrics.com} questionnaire. The participants completed background measures on demographics and personality, and then proceeded to the main part of the survey where they read a random selection of five of the news articles. 
After reading each of the articles, participants were asked to rate their level of empathy and distress before describing their thoughts and feelings about it in writing. 

In contrast to previous work, this set-up allowed us to acquire empathy scores of the actual {\it writer} of a text, instead of having to rely on an external evaluation by third parties (often student assistants with background in computer science). Arguably, our proposed annotation methodology yields more appropriate gold data, yet also leads to more variance in the relationship between linguistic features 
and empathic state ratings. That is because each rating reflects a single individual's feelings rather than a more stable average assessment by multiple raters. To account for this, we use {\it multi-item scales} as is common practice in psychology. I.e., participants give ratings for multiple items measuring the same construct (e.g., empathy) which are then averaged to obtain more reliable results. 
%
As far as we know, this is the first time that multi-item scales are used in sentiment analysis.\footnote{ Here, we use \textit{sentiment} as an umbrella term subsuming semantic orientation, emotion, as well as highly related concepts such as empathy.}

In our case, participants used Batson's Empathic Concern -- Personal Distress Scale (see above), i.e, rating 6 items for empathy (e.g., {\it warm, tender, moved}) and 8 items for  distress (e.g., {\it troubled, disturbed, alarmed}) using a 7-point scale for each of those (see Appendix for details).
%
%
After rating their empathy, participants were asked to share their feelings about the article as they would with a friend in a private message or with a group of friends as a social media post in 300 to 800 characters. 
Our final gold standard consists of these {\it messages} combined with the numeric ratings for empathy and distress. 

In sum,  403 participants completed the survey. Median completion time was 32 minutes and each participant received 4 USD as compensation. 
%





\textbf{Post-Processing. }
Each message was manually reviewed by the authors. Responses which  deviated from the task description (e.g., mere copying from the articles at display) were removed (31 responses, 155 messages), leading to a total 1860 messages in our final corpus. Gold ratings for empathy and distress were derived by averaging the respective items of the two multi-item scales. 


\section{Corpus Analysis}
\label{sec:analysis}

For a first impression of the language of our new  gold standard, we provide illustrative examples in Table \ref{tab:examples}. The participant in Example (1) displays higher empathy than distress, (2) displays higher distress than empathy, and (3) shows neither empathic state, but employs sarcasm, colloquialisms and social-media-style acronyms to express lack of emotional response to the article. As can be seen, the language of our corpus is diverse and authentic, featuring many phenomena of natural language which render its computational understanding difficult, thus constituting a sound but challenging gold standard for empathy prediction.

{\bf Token Counts.} We tokenized the 1860 messages using NLTK tools \cite{Bird06}. In total, our corpus amounts to $173,686$ tokens. Individual message length varies between 52 and 198 tokens, the median being 84. 
See Appendix for details.


{\bf Rating Distribution. } 
\begin{figure}[t]
    \centering
    \includegraphics[width=.5\textwidth]{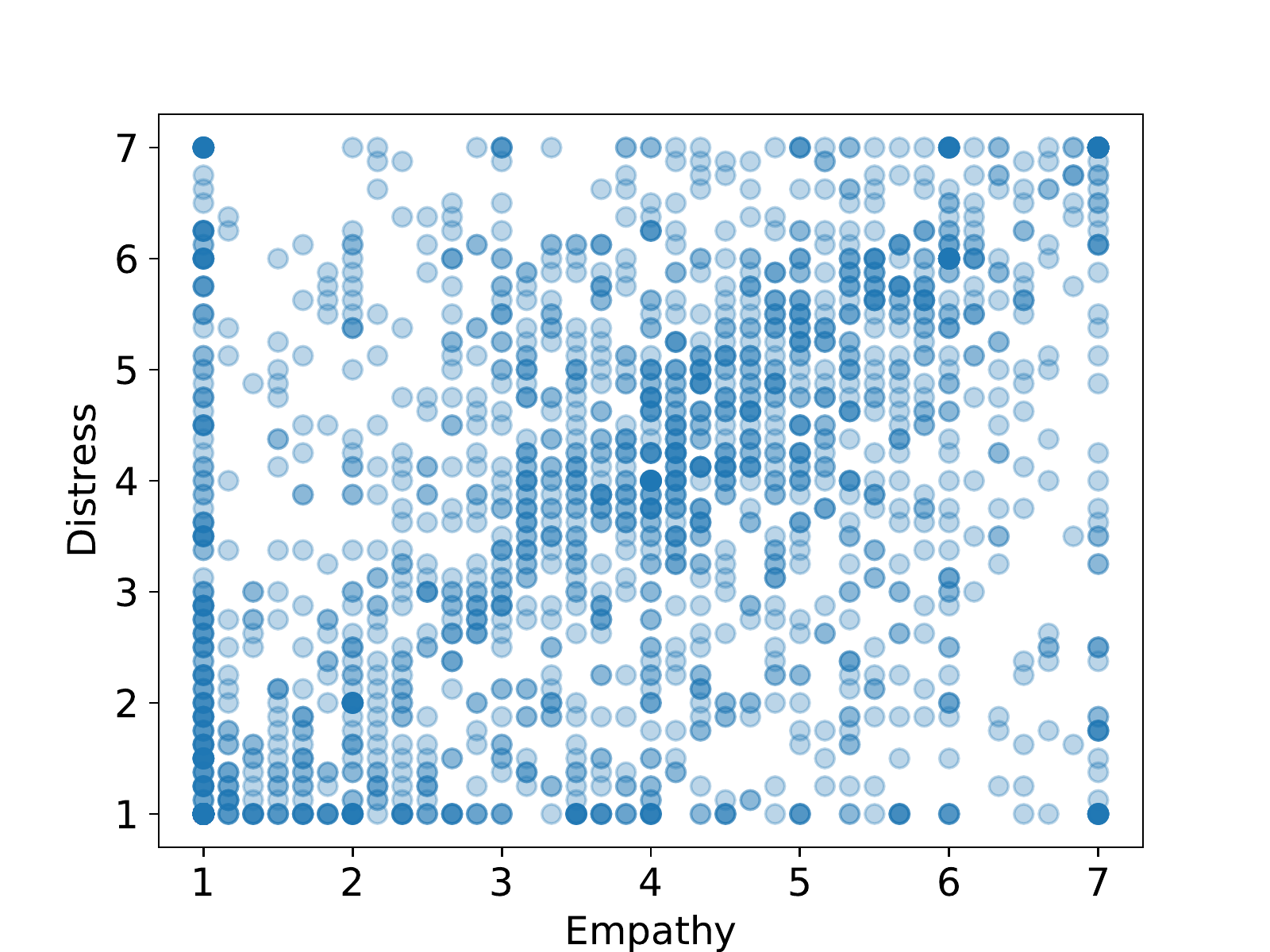}
    \vspace*{-10pt}
    \caption{Scatter plot of the bivariate distribution of empathy and distress ratings.}
    \label{fig:empathy_vs_distress}
\end{figure}
Figure \ref{fig:empathy_vs_distress} displays the bivariate distribution of empathy and distress ratings. As can be seen both target variables have a clear linear dependence, yet show only a moderate Pearson correlation of $r{=}.451$, similar to what was found in prior research \cite{batson1987distress,batson1997empathy}.
This finding supports that the two scales capture distinct affective phenomena and underscores the importance of our decision to describe empathic states in terms of {\it multiple} target variables, constituting a clear advancement over previous work. 
Both kinds of ratings show good coverage over the full range of the scales.



{\bf Reliability of Ratings. } Since each message is annotated by only one rater, its author, typical measures of inter-rater agreement are not applicable. 
Instead, we compute {\it split-half reliability} (SHR), a standard approach in psychology \cite{cronbach1947test} which also becomes increasingly popular in sentiment analysis \cite{Mohammad17wassa,Buechel18coling}. SHR is computed by splitting the ratings for the individual scale items (e.g., {\it warm}, {\it tender}, etc. for empathy) of all participants randomly into two groups, averaging the individual item ratings for each group and participant, and then measuring the  correlation between both groups. This process is repeated 100 times with random splits, before again averaging the results.
Doing so for empathy and distress, we find very high\footnote{
     For a comparison against previously reported SHR values for different emotional categories, see \newcite{Mohammad17starsem}.
} 
SHR values of $r{=}.875$ and $.924$, respectively.

\section{Modeling Empathy and Distress}
\label{sec:modeling}

In this section, we provide experimental results for modeling empathy and distress ratings based on the participants' messages (see Section \ref{sec:corpus}). We examine three different types of models, varying in design complexity. Distinct models were trained for empathy and distress prediction.

First, ten percent of our newly created gold standard were randomly sampled to be used in development experiments. Then, the main experiment was conducted using 10-fold cross-validation (CV), providing each model with identical train-test splits to increase reliability. The dev set was excluded for the CV experiment. 

Model performance is measured in terms of Pearson correlation $r$ between predicted values and the human gold ratings. Thus, we phrase the prediction of empathy and distress as regression problems.

The input to our models 
is based on word embeddings, namely the publicly available FastText embeddings 
which were trained on Common Crawl (${\approx}600$B tokens) \cite{Bojanowski17,Mikolov18advances}.

{\bf Ridge. } Our first approach is Ridge regression, an $\ell^2$-regularized version of linear regression. The centroid of the word embeddings of the words in a message is used as features (embedding centroid). The regularization coefficient $\alpha$ is automatically chosen  from $\{1, .5, .1, ..., .0001\}$ during training.

{\bf FFN. } Our second approach is a Feed-Forward Net with two hidden layers (256 and 128 units, respectively) with ReLU activation. Again, the embedding centroid is used as features.

{\bf CNN. } The last approach is a Convolutional Neural Net.\footnote{
   Recurrent models did not perform well during development due to high sequence length.
    } 
We use a single convolutional layer with filter sizes 1 to 3, each with 100 output channels, followed by an average pooling layer and a dense layer of 128 units. ReLUs were used for the convolutional and again for the dense layer.

Both deep learning models were trained using the Adam optimizer \cite{Kingma15} with a fixed learning rate of $10^{-3}$ and a batch size of 32. We trained for a maximum of 200 epochs yet applied early stopping if the performance on the validation set did not improve for 20 consecutive epochs. We applied dropout with probabilities of $.2$, $.5$ and $.5$ on input, dense and pooling layers, respectively. Moreover $\ell^2$ regularization of $.001$ was applied to the weights of conv and dense layers. Word embeddings were not updated.
\begin{table}[t]
    \centering
    \small
\begin{tabular}{lrrr}
\toprule
 & {\bf Empathy} & {\bf Distress} & {\bf Mean} \\
\midrule
Ridge & .385\phantom{*} & .410\phantom{*} & .398\phantom{*}\\
FFN & .379\phantom{*} & .401\phantom{*} & .390\phantom{*}\\
CNN & {\bf.404}* & {\bf .444}* & {\bf.424}*\\
\bottomrule
\end{tabular}
    \caption{Model performance for predicting empathy and distress in Pearson's $r$; with row-wise mean; best result per column in bold, significant ($p<.05$) improvement over other models marked with `*'.}
    \label{tab:performance}
\end{table}

The results are provided in Table \ref{tab:performance}. As can be seen, all of our models achieve satisfying performance figures ranging between $r{=}.379$ and $.444$, given the assumed difficulty of the task (see Section \ref{sec:analysis}). 
On average over the two target variables, the CNN performs best, followed by Ridge and the FFN. While the CNN significantly outperforms the other models in every case, the differences between Ridge and the FFN are not statistically significant for either empathy or distress.\footnote{We use a two-tailed $t$-test for paired samples based on the results of the individual CV runs; $p<.05$.}
The improvements of the CNN over the other two approaches are much more pronounced for distress than for empathy. 
Since only the CNN is able to capture semantic effects from composition and word order, our data suggest that these phenomena are more important for predicting distress, whereas lexical features alone already perform quite well for empathy.

{\bf Discussion. } 
In comparison to closely related tasks such as emotion prediction \cite{Mohammad17wassa} our performance figures for empathy and distress prediction are generally lower. However, given the small amount of previous work for the problem at hand, we argue that our results are actually quite strong. This becomes obvious, again, in comparison with emotion analysis where early work achieved correlation values around $r{=}.3$ at most \cite{Strapparava07}. Yet state-of-the-art performance literally doubled over the last decade \cite{Beck17}, in part due to much larger training sets.

Comparison to the limited body of previous work in text-based empathy prediction  is difficult for a number of reasons, e.g., differences in domain, evaluation metric, as well as methodology and linguistic level of annotation.
\newcite{Khanpour17} annotate and model empathy  in online health communities on the \textit{sentence}-level, whereas the instances in our corpus are much longer and comprise multiple sentences. In contrast to our work, they treat empathy prediction as a classification problem. Their best performing model, a CNN-LSTM, achieves an F-score of .78.
\newcite{Gibson15} predict therapists' empathy in motivational interviews. Each therapy session transcript received one numeric score. Thus, each prediction is based on much more language data than our individual messages comprise.
%
Their best model achieves a Spearman rank correlation of $.61$ using $n$-gram and psycholinguistic features.

Our contribution goes beyond both of these studies by, first, enriching empathy prediction with personal distress and, second, by annotating and modeling the empathic state actually felt by the writer, instead of relying on external assessments. 

\section{Conclusion}
This contribution was the first to attempt empathy prediction in terms of {\it multiple} target variables, empathic concern and personal distress.
We proposed a novel annotation methodology 
capturing empathic states actually felt by the author of a statement, instead of relying on third-party assessments. 
To ensure high reliability in this single-rating setting, we employ multi-item scales in line with best practices in psychology. 
Hereby we create the first publicly available gold standard for empathy prediction in written language, our survey being set-up and supervised by an expert psychologist.
Our analysis shows that the data set excels with high rating reliability and an authentic and diverse language, rich of challenging phenomena such as sarcasm. We provide experimental results for three different predictive models, our CNN turning out superior.


\section*{Acknowledgments}
Sven Buechel would like to thank his doctoral advisor Udo Hahn, JULIE Lab, for funding his research visit at the University of Pennsylvania.

\bibliography{literature}
\bibliographystyle{stylesheet/acl_natbib_nourl}

\appendix
\section{Supplemental Material}
\label{sec:supplemental}

\subsection*{Details on Stimulus and Instructions}

Before being used in our survey, the selected news articles were categorized by the research interns who gathered them in terms of their intensity of suffering (major or minor), cause of suffering (political, human, nature or other), patient of suffering (humans, animals, environment, or other) and scale of suffering (individual or mass). Research interns also provided a short list of key words for each article. This additional information was gathered to examine the influence of these factors on empathy elicitation and modeling performance in later studies.

At the beginning of the survey participants completed background items covering general demographics (including age, gender, and ethnicity),  the most commonly used {\it trait} empathy scale, the Interpersonal Reactivity Index \cite{davis1980interpersonal}, a brief assessment of the Big 5 personality traits  \cite{gosling2003very}, life satisfaction \cite{diener1985satisfaction}, as well as a brief measure of generalized trust.

\begin{figure}[ht]
    \centering
    \fbox{
    \includegraphics[width=.45\textwidth]{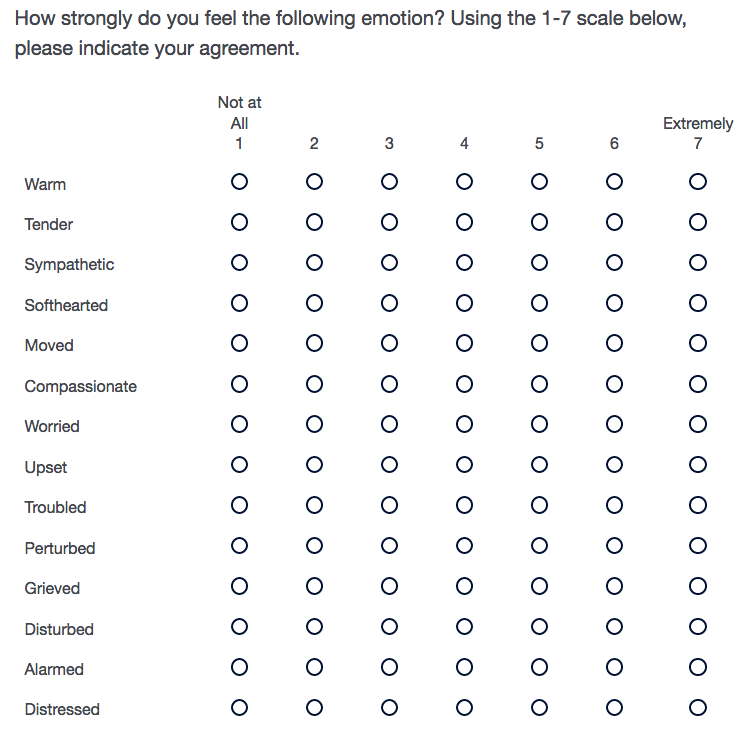}
    }
    \caption{Multi-item scales for empathic concern and personal distress.}
    \label{fig:scales}
\end{figure}

After reading each of the articles, participants rated their level of empathic concern and personal distress using multi-item scales. {\bf Figure \ref{fig:scales}} shows a cropped screenshot of the survey hosted on \texttt{Qualtrics.com}. 
The first six items ({\it warm, tender, sympathetic, softhearted, moved}, and {\it compassionate}) refer to empathy. The last eight items ({\it worried, upset, troubled, perturbed, grieved, disturbed, alarmed}, and {\it distressed}) refer to distress.

After completing the rating items, participants were instructed to describe their reactions in writing as follows:
{\it  Now that you have read this article, please write a message to a friend or friends about your feelings and thoughts regarding the article you just read. This could be a private message to a friend or something you would post on social media. Please do not identify your intended friend(s) --- just write your thoughts about the article as if you were communicating with them. Please use between 300 and 800 characters.}

\subsection*{Further Corpus Analyses}

\begin{figure}[ht]
    \centering
    \vspace*{-0pt}
    \includegraphics[width=.4\textwidth]{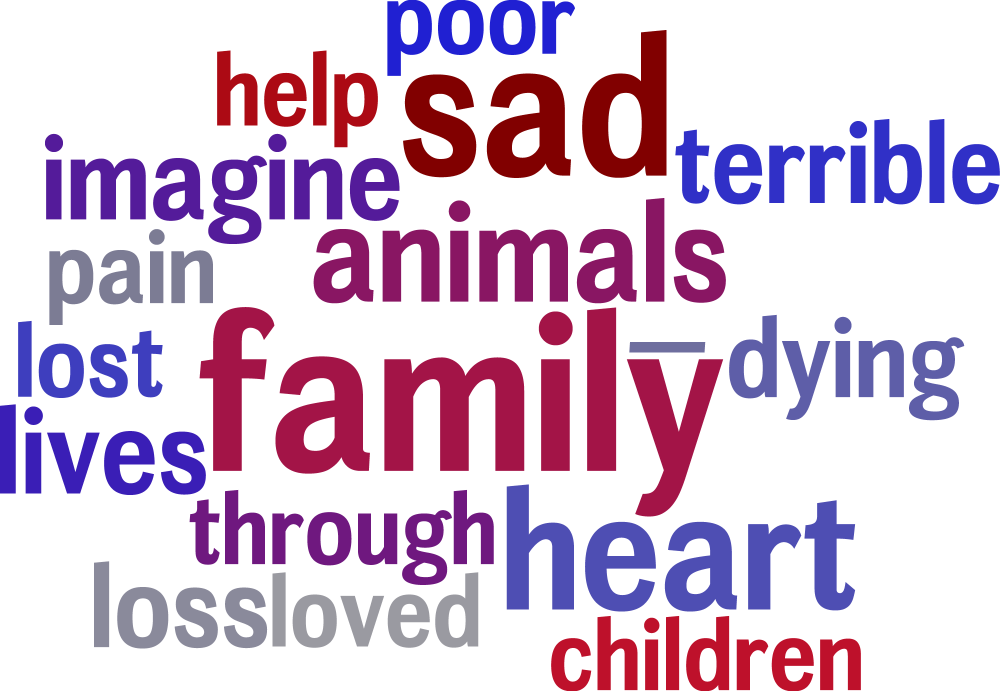}
    \caption{Word cloud of high empathy 1-grams.}
    \label{fig:wordcloud_empathy}
\end{figure}

\begin{figure}[ht]
    \centering
    \vspace*{-0pt}
    \includegraphics[width=.4\textwidth]{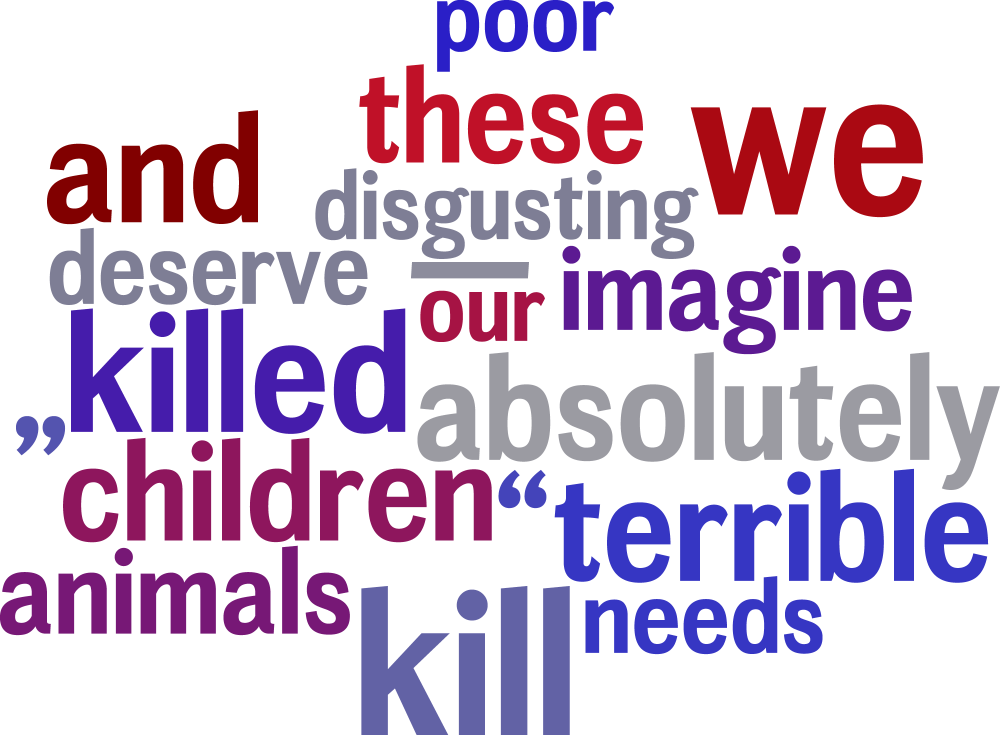}
    \caption{Word cloud of high distress 1-grams.}
    \label{fig:wordcloud_distress}
\end{figure}

The word clouds in {\bf Figure  \ref{fig:wordcloud_empathy}} and  {\bf Figure \ref{fig:wordcloud_distress}} show 1-grams of our corpus which correlate significantly (Benjamini-Hochberg corrected $p < .05$) with high empathy and high distress ratings, respectively. In the word clouds, larger size indicates higher correlation and the color scale, gray-blue-red, indicates word frequency, dark red being most prevalent. The Differential Language Analysis Toolkit \cite{schwartz2017dlatk} was utilized for this analysis.
As can be seen, the word clouds display high face-validity, giving further evidence for the soundness of our acquisition methodology.


\begin{figure}[ht]
    \centering
    \vspace*{-0pt}
    \includegraphics[width=.5\textwidth]{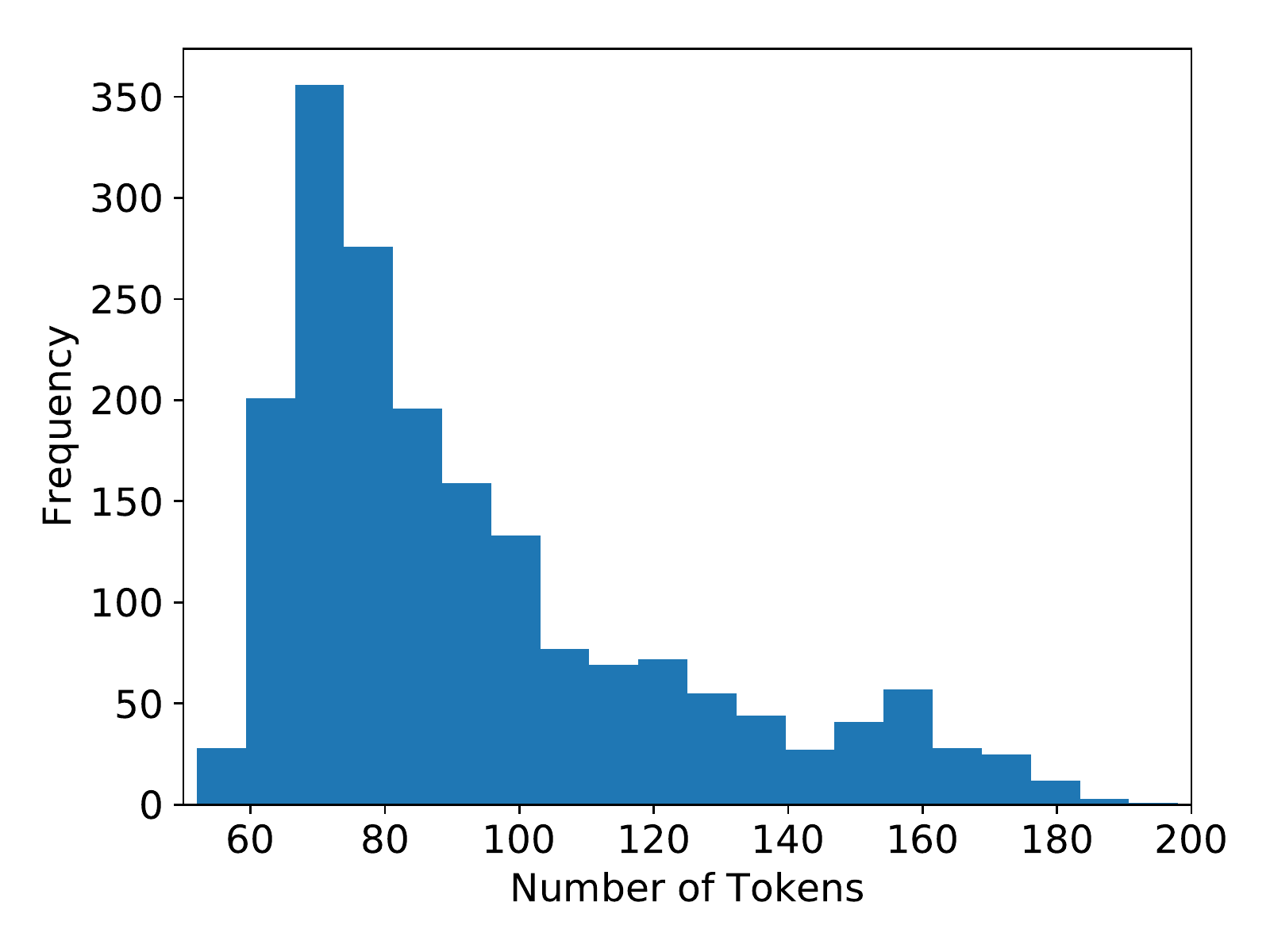}
    \vspace*{-12pt}
    \caption{Histogram of message length in our corpus.}
    \label{fig:hist_token_counts}
    \vspace*{-12pt}
\end{figure}

{\bf Figure \ref{fig:hist_token_counts}} displays the distribution of the message length of our corpus in tokens. As can be seen the majority of messages contain between 60 and 100 tokens. Yet outliers go up to almost 200. The introduction of a character cap for the writing task proved successful in comparison to a pilot study where this measure has not been in place. In the latter case, the maximum number of tokens was nearly twice as high due to even stronger outliers.

\end{document}